\documentclass[conference]{IEEEtran}
\IEEEoverridecommandlockouts
\usepackage[T1]{fontenc}
\usepackage[utf8]{inputenc}
\usepackage{times}
\usepackage{latexsym}
\usepackage{microtype}

\usepackage{amsmath,amssymb,amsfonts}

\usepackage{cite}
\usepackage{url}

\usepackage{graphicx}
\usepackage[table]{xcolor} % Load xcolor once with [table] if needed
\usepackage{textcomp}
\usepackage{booktabs}
\usepackage{multirow}
\usepackage{multicol}
\usepackage{tabularx}
\usepackage{adjustbox}
\usepackage[normalem]{ulem} % Load after xcolor to avoid clashes

\usepackage{algorithmic}

\usepackage{tikz}
\usepackage{forest}
\usetikzlibrary{trees,positioning,shapes,shadows,arrows.meta}

\usepackage{pgf}
\usepackage{pgfplots}
\pgfplotsset{compat=newest, width=\textwidth}

\usepackage{collcell}

\usepackage{caption}

\pgfplotsset{compat=newest, width=\textwidth}

\newcommand{\CellColor}[1]{%
  \pgfmathparse{int(round(#1))} % Calculate the intensity based on the value
  \ifnum\pgfmathresult>100
    \def\temp{100}
  \else
    \ifnum\pgfmathresult<0
      \def\temp{0}
    \else
      \def\temp{\pgfmathresult}
    \fi
  \fi
  \pgfmathsetmacro{\CellColorValue}{\temp}
  \edef\x{\noexpand\cellcolor{teal!\CellColorValue!white}}\x #1
}

\newcommand{\CellColorA}[1]{%
  \pgfmathsetmacro{\ColorValue}{100*(#1-1)/29} % Scale from 1 to 30
  \edef\x{\noexpand\cellcolor{teal!\ColorValue!white}}\x #1
}

\newcommand{\CellColorB}[1]{%
  \pgfmathsetmacro{\ColorValue}{100*#1} % Scale from 0 to 1
  \edef\x{\noexpand\cellcolor{teal!\ColorValue!white}}\x #1
}

\newcommand{\CheckmarkCell}{\cellcolor{darkgreen}\checkmark}
\newcommand{\CrossCell}{\cellcolor{red}$\times$}

\def\BibTeX{{\rm B\kern-.05em{\sc i\kern-.025em b}\kern-.08em
    T\kern-.1667em\lower.7ex\hbox{E}\kern-.125emX}}

\begin{document}

\title{Code LLMs: A Taxonomy-based Survey}

\author{\IEEEauthorblockN{Nishat Raihan}
\IEEEauthorblockA{\textit{George Mason University}\\
Fairfax, VA, USA \\
mraihan2@gmu.edu}
\and
\IEEEauthorblockN{Christian Newman}
\IEEEauthorblockA{\textit{Rochester Institute of Technology}\\
Rochester, NY, USA \\
cdnvse@rit.edu}
\and
\IEEEauthorblockN{Marcos Zampieri}
\IEEEauthorblockA{\textit{George Mason University}\\
Fairfax, VA, USA \\
mzampier@gmu.edu}
}

\maketitle

\begin{abstract}
Large language models (LLMs) have demonstrated remarkable capabilities across various NLP tasks and have recently expanded their impact to coding tasks, bridging the gap between natural languages (NL) and programming languages (PL). This taxonomy-based survey provides a comprehensive analysis of LLMs in the NL-PL domain, investigating how these models are utilized in coding tasks and examining their methodologies, architectures, and training processes. We propose a taxonomy-based framework that categorizes relevant concepts, providing a unified classification system to facilitate a deeper understanding of this rapidly evolving field. This survey offers insights into the current state and future directions of LLMs in coding tasks, including their applications and limitations. 
\end{abstract}


\section{Introduction}

LLMs pretrained on very large corpora \cite{radford2018improving}, show adaptability to various NLP tasks, including challenging \textit{NL-PL} tasks. Models built on top of them, such as CodeBERT \cite{feng2020codebert} and GraphCodeBERT \cite{guo2020graphcodebert} have been instrumental in bridging human and computational language processing. 

Causally masked, decoder-only language models such as GPT \cite{radford2018improving} are introduced before encoder-only models like BERT \cite{devlin2018bert} or Electra \cite{clark2020electra}. However, they garner significant attention with the release of OpenAI's GPT-2 \cite{radford2019language} and subsequent GPT-3 \cite{brown2020language} models, demonstrating efficiency in a wide range of NLP tasks, including zero-shot learning and programming language comprehension. OpenAI also introduces HumanEval \cite{brown2020language} - a benchmark for code generation. Despite their capabilities, these models are proprietary, restricting further experimentation by the research community. In response, open-source alternatives like Falcon \cite{almazrouei2023falcon} and MetaAI's LLaMA \cite{touvron2023llama} are developed. They leverage sophisticated training techniques such as Rotary Positional Embeddings \cite{su2024roformer}, Grouped Multi Query Attention, and the SwiGLU activation function \cite{shazeer2020glu}, aiming to offset their parameter deficit. However, they typically underperform compared to their closed-source counterparts like GPT-4 and Anthropic's Claude \cite{claude2023} on coding benchmarks - most likely due to the use of public corpora like Wikipedia \cite{WikipediaCorpus} and Common Crawl \cite{CommonCrawlCorpus}, which lack a rich representation of programming languages.

\begin{figure*} [!h]
    \centering
    \includegraphics[width=.9\linewidth]{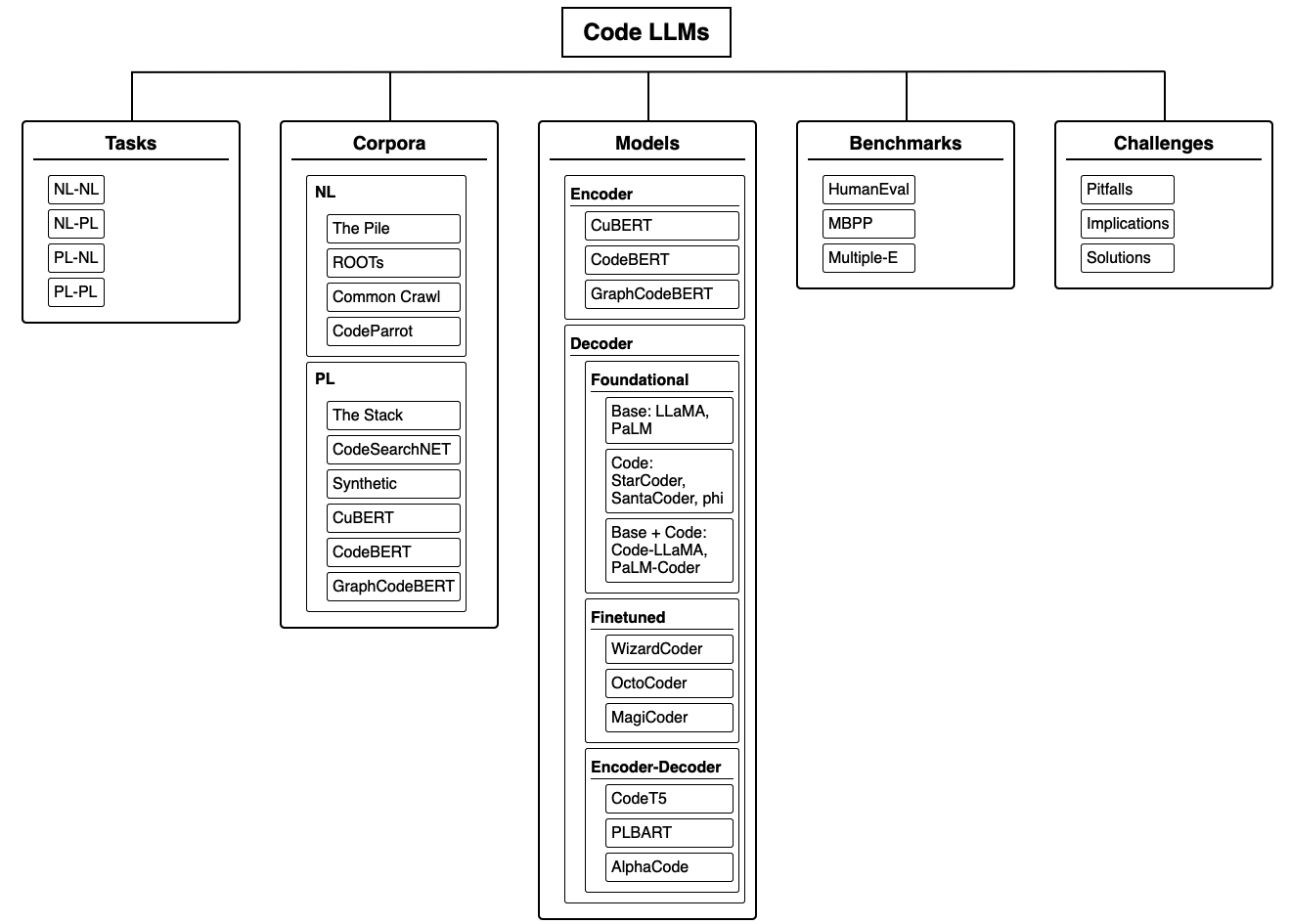}
    \caption{Code LLMs - Area Taxonomy.}
    \label{fig:taxonomy}
\end{figure*}

The contributions of this paper are as follows. 

\begin{itemize}
    \item We present the first taxonomy-based survey on Code LLMs.
    \item We examine decoder-based models as well as encoder and encoder-decoder architectures, providing in-depth analysis of model structures and nuances.
    \item We discuss relevant benchmarks and corpora, addressing potential challenges and their solutions.
    \item We provide insights on open problems in the field in order to move forward.
\end{itemize}

The remainder of this paper is structured as follows: Section II presents a comprehensive taxonomy of the field, encompassing tasks, datasets, models, benchmarks, and challenges, to provide a holistic view. Section III offers a critical analysis of the literature, comparing and contrasting different approaches. Section IV focuses on unresolved issues and open problems within the domain, pointing to directions for future research. Throughout these sections, we will consistently refer back to our primary objectives, ensuring that our analysis provides clear insights into the educational challenges \cite{raihan2024large}, ethical considerations, and performance trends of LLMs in coding tasks.

\section{Area Taxonomy}

Figure \ref{fig:taxonomy} presents a comprehensive taxonomy of research in the area of Code Large Language Models (Code LLMs). The taxonomy organizes the field into five principal sub-areas:

\begin{enumerate}
    \item \textbf{Tasks}: We classify coding tasks into four distinct categories based on the nature of their inputs and outputs, which may involve Natural Language (NL) and/or Programming Language (PL). These categories are:
    \begin{itemize}
        \item \textbf{NL-NL}: Tasks where both input and output are in natural language.
        \item \textbf{NL-PL}: Tasks where input is in natural language, and output is in a programming language.
        \item \textbf{PL-PL}: Tasks where both input and output are in a programming language.
        \item \textbf{PL-NL}: Tasks where input is in a programming language, and output is in natural language.
    \end{itemize}
    This classification provides a systematic framework to organize and focus on various coding tasks effectively. Our survey emphasizes Code LLMs' performance in generative tasks (NL-PL), particularly code generation from human language prompts.

    \item \textbf{Corpora}: The success of transformer-based models hinges on pretraining with large, diverse datasets. For Code LLMs, corpora are categorized into:
    \begin{itemize}
        \item \textbf{PL-corpora}: Datasets composed of programming languages.
        \item \textbf{NL-corpora}: Datasets consisting of natural language text.
    \end{itemize}
    Recent efforts have focused on compiling specialized programming corpora to enhance the models' understanding of coding tasks.

    \item \textbf{Models}: We analyze three primary types of Transformer-based models \cite{vaswani2017attention}:
    \begin{itemize}
        \item \textbf{Encoder-only models}: Typically used for understanding tasks.
        \item \textbf{Decoder-only models}: Primarily effective for generative tasks, which are the focus of this survey.
        \item \textbf{Encoder-decoder models}: Employed for tasks requiring both understanding and generation.
    \end{itemize}
    Decoder-only models are further categorized into:
    \begin{itemize}
        \item \textbf{Foundational models}: Pretrained on extensive corpora and categorized based on their data sources as \textit{base}, \textit{code}, or \textit{mixed}.
        \item \textbf{Fine-tuned models}: Adapted for specific tasks to improve performance.
    \end{itemize}

    \item \textbf{Benchmarks}: The proliferation of Code LLMs necessitates robust benchmarks to evaluate their performance. These benchmarks serve as critical tools for comparing models and assessing their capabilities across various tasks.

    \item \textbf{Challenges}: Finally, we highlight the key challenges in this domain, including:
    \begin{itemize}
        \item Potential limitations in existing models and corpora.
        \item Ethical concerns surrounding the use of generated code.
        \item Real-world applicability and the need for more interpretable, reliable models.
    \end{itemize}
    We also propose directions for addressing these challenges to ensure sustainable advancements in the field.
\end{enumerate}

This taxonomy provides a structured framework for navigating the research landscape, highlighting interconnections between sub-areas, and fostering a systematic approach to future research and development in Code LLMs.

\section{Taxonomy-based Survey}
Now we take a deeper dive into each of these subdomains and analyze them further.

\subsection{\textbf{Tasks}} \label{tasks}

The spectrum of relevant tasks in the domain can be categorized into four main types: NL-NL, PL-NL, PL-PL, and NL-PL. While each category serves distinct purposes, our primary focus lies on NL-PL tasks, particularly code generation, which showcases the primary purpose of decoder-based Code LLMs.

Earlier works include NL-NL tasks mainly, such as document translation and log analysis, traditionally driven by standard NLP techniques. As code-specific applications gained prominence, PL-NL tasks like code summarization, comment generation, and identifier normalization emerged \cite{haiduc2010use, mcmillan2011exemplar, poshyvanyk2013using}, bridging the gap between code and human understanding. Early approaches used statistical methods, but neural networks now offer greater accuracy. Identifier normalization, including abbreviation expansion and variable name splitting \cite{lawrie2010normalizing, corazza2012linsen, lawrie2011expanding}, is essential for improving code comprehension and quality. Latter works explore PL-PL tasks focus on refining existing code, including translation, completion, infilling, refactoring \cite{fowler1999refactoring}, and code smell detection \cite{marinescu2004detection}. Models such as CodeBERT \cite{feng2020codebert} and GraphCodeBERT \cite{guo2020graphcodebert} have excelled in tasks like code infilling by framing it as a masked language modeling problem.

Current works focus on advanced NL-PL tasks, notably code generation, where models must interpret natural language instructions and programming concepts, including syntax, semantics, and frameworks, to produce executable code. These models are evaluated by how accurately generated code passes test cases. Recent advances in autoregressive models, particularly GPT-4 \cite{openai2023gpt4}, have improved performance across all task types, though challenges remain in complex areas like code translation and ensuring the correctness of generated code. As research progresses, Code LLMs continue to enhance automation and augment software development processes.

\begin{table}[!h]
\centering
\begin{tabular}{@{}ccccc@{}}
\toprule
Type & Name & Size & PLs & Source \\ \midrule
         & Common Crawl   & 390 TB   & --   & Web   \\
NL       & ROOTs   & 1.6 TB   & --   & Web   \\
         & The Pile  & 886 GB   & --   & Web   \\ \midrule
    & The Stack [v2]  & 12.4 TB   & 619   & Github \& Kaggle  \\
   & The Stack [v1]  & 3.1 TB   & 30   & Github  \\
PL       & CodeParrot   & 1 TB   & 30   & Github   \\
   & CodeSearchNet   &  3.6 GB  & 6   & Github   \\
   & Synthetic [phi-1]   &  1B tokens  &  1  & LLMs   \\ \bottomrule
\end{tabular}
\caption{Common Corpora used for Pre-training Code-LLMs }
\label{table1:corpora}
\end{table}

\subsection{\textbf{Corpora}}
General purpose pretrained models get trained on large corpora of text which often lack a rich distribution of programming languages. Also, these corpora filter out code-related data, which enables superior natural language understanding but often results in suboptimal performance on coding tasks. To improve competency in the programming domain, some code-focused corpora have been compiled in recent years as well. These are compiled using publicly available code repositories like GitHub\footnote{\url{www.github.com}} or Kaggle\footnote{\url{www.kaggle.com}}. Most recently, some of the LLMs also use synthetic corpora generated by other LLMs, as shown in Table \ref{table1:corpora}. The Code LLMs incorporate training on these code-heavy resources along with general-purpose corpora. As shown in Table \ref{table1:corpora}, we categorize the most used training corpora into two subdomains: NL and PL.

\subsection{\textbf{Models}}

Building on transformer architectures \cite{vaswani2017attention}, several works have explored encoder-only (CodeBERT \cite{feng2020codebert}), encoder-decoder (CodeT5 \cite{wang2021codet5}), and decoder-only variants. Currently, decoder-only models like GPT4 \cite{openai2023gpt4} and CodeLLaMA \cite{roziere2023code} achieve state-of-the-art results in code generation tasks. We provide an in-depth analysis for each of these model types.

\vspace{.15cm}

\subsubsection{\textbf{Encoders}} The foundational encoder-only model, BERT \cite{devlin2018bert}, uses masked token prediction and next sentence prediction as the pretraining tasks. They evolved to be just masked span detection (ERNIE \cite{zhang2019ernie}) and corrupted token prediction (ELECTRA \cite{clark2020electra}) for efficiency gains. As shown in Table \ref{tab:masked-type}, iterative innovations in masking objectives have cumulatively enhanced representation learning and downstream performance in coding tasks.

% Define colors (if not predefined)
\definecolor{darkgreen}{rgb}{0.0, 0.99, 0.5}
\definecolor{red}{rgb}{0.99, 0.4, 0.4}

\begin{table}[ht]
\centering
\begin{tabular}{@{}ccccc@{}}
\toprule
& \multicolumn{4}{c}{Type of Masked Tokens} \\
\midrule
Model & NL & PL & AST & CFG \\
\midrule
BERT \cite{devlin2018bert} & \CheckmarkCell & \CrossCell & \CrossCell & \CrossCell \\
RoBERTa \cite{liu2021robustly} & \CheckmarkCell & \CrossCell & \CrossCell & \CrossCell \\ \midrule
CodeBERT \cite{feng2020codebert} & \CheckmarkCell & \CheckmarkCell & \CrossCell & \CrossCell \\
CuBERT \cite{kanade2020learning} & \CheckmarkCell & \CheckmarkCell & \CrossCell & \CrossCell \\
TreeBERT \cite{jiang2021treebert} & \CrossCell & \CheckmarkCell & \CheckmarkCell & \CrossCell \\
GraphCodeBERT \cite{guo2020graphcodebert} & \CheckmarkCell & \CheckmarkCell & \CheckmarkCell & \CrossCell \\
SyncoBERT \cite{wang2021syncobert} & \CheckmarkCell & \CheckmarkCell & \CheckmarkCell & \CrossCell \\
Code-MVP \cite{wang2022code} & \CheckmarkCell & \CheckmarkCell & \CheckmarkCell & \CheckmarkCell \\
\bottomrule
\end{tabular}
\caption{Masked Language Models [Encoders] and the type of Masked Tokens during their pre-training phase.}
\label{tab:masked-type}
\end{table}

\begin{table*}[!t]
\centering
\begin{tabular}{@{}c|cccccc@{}}
\toprule
Task & Code-to-Doc & Code-to-NL  & Summarization & Search & Clone Det. & Defect Det. \\
BenchMarks  & CodeSearchNet \cite{husain2019codesearchnet} & CodeNN (C\#) \cite{CodeNN} & ETH Py150 \cite{Python150kDataset} & CoSQA \cite{huang2021cosqa} & BigCloneBench \cite{svajlenko2014towards} & GREAT \cite{hellendoorn2019global} \\ 
Acc. Metric  & BLEU-4 & BLEU & prec & MRR & F1 & acc \\ 
\midrule
\midrule
RoBERTa \cite{liu2021robustly} & 16.57 & 19.81 & $\times$ & 0.58 & 0.94 & 0.82 \\ 
CuBERT \cite{kanade2020learning}  & 17.41 & 14.95 & 0.46 & $\times$ & $\times$ & $\times$ \\
CodeBERT \cite{feng2020codebert}  & 17.83 & \textbf{22.36} & 0.49 & 0.64 & 0.94 & 0.86 \\
TreeBERT \cite{jiang2021treebert}  & \textbf{20.49} & 17.94 & \textbf{0.61} & $\times$ & $\times$ & $\times$ \\
GraphCodeBERT \cite{guo2020graphcodebert}  & $\times$ & $\times$ & $\times$ & 0.68 & 0.95 & 0.88 \\
SynCoBERT \cite{wang2021syncobert}  & $\times$ & $\times$ & $\times$ & 0.70 & \textbf{0.97} & 0.88 \\
Code-MVP \cite{wang2022code}  & $\times$ & $\times$ & $\times$ & \textbf{0.72} & $\times$ & \textbf{0.89} \\
\bottomrule
\end{tabular}
\caption{Performance of Encoder-only models on six common tasks. The results mentioned in the table are for tasks where models use the same benchmark dataset and report the results using the same accuracy metric. }
\label{tab:enc-perform}
\end{table*}

Earlier encoders such as CuBERT \cite{kanade2020learning} use unified pretraining (NL-NL or PL-PL pairs) and MLM as the pretraining task. While CodeBERT \cite{feng2020codebert} focuses on both MLM and corrupted token prediction. During the pretraining phase, it uses both unimodal (NL-NL or PL-PL pairs) and bimodal (NL-PL pairs) to have a better understanding of the coding contexts. Latter models include TreeBERT \cite{jiang2021treebert} that modifies the MLM task as a Tree MLM (TMLM) task, where instead of predicting missing NL or PL tokens, the model tries to predict the missing nodes of the Abstract Syntax Trees (ASTs). ASTs are generated using open source tools like \textit{tree-sitter}\footnote{\url{https://github.com/tree-sitter/tree-sitter}}. Combining these methods, GraphCodeBERT \cite{guo2020graphcodebert} includes the ASTs along with the NL-PL sentence pair to provide more structural context for the code. This trio-MLM structure is further improved by SynCoBERT \cite{wang2021syncobert}, with two newly introduced pre-training tasks called Identifier Prediction, where the models attempt to predict which tokens from the source code are identifiers (\textit{Identifiers in source code refer to the names given to programmatic elements like variables, functions, etc. which are used to represent and reference them within code.}) and AST Edge Prediction, where the model predicts the edges in the ASTs provided. Further improved by Code-MVP \cite{wang2022code} with the inclusion of Control Flow Graphs (CFGS) along with NL-PL-AST trio. CFGs are constructed using publicly available tools like \textit{Scalpel} \cite{li2022scalpel}. In Table \ref{tab:masked-type}, we list the types of masking these models employ.

Encoder-only models excel in tasks like documentation generation, code search, summarization, and defect detection, but comparing their performance is challenging due to inconsistent reporting across tasks, evaluation metrics, and datasets. In Table \ref{tab:enc-perform}, we focus on benchmarks with the broadest reporting, using a uniform accuracy metric.

\subsubsection{\textbf{Encoder-Decoders}} These models employ a full transformer architecture like T5 \cite{raffel2020exploring} and BART \cite{lewis2020bart}.
Building on T5, models like PyMT5 \cite{clement2020pymt5} and CodeT5 \cite{wang2021codet5} have been proposed for coding tasks in Python. CodeT5 incorporates identifier prediction and finetuning on tasks like code generation and summarization. Latter works like CodeT5+ \cite{wang2023codet5+} and SPT-Code \cite{niu2022spt} perform two-stage pretraining, first on unimodal then bimodal data, adding ASTs for better understanding. Deepmind's AlphaCode \cite{li2022competition} achieves state-of-the-art code generation via larger models, scaling up to 41B parameters, pretrained on high-quality GitHub data. Other works build on top of BART \cite{lewis2020bart}, such as PLBART \cite{ahmad2021unified} which unifies unimodal and bimodal pretraining. Unlike encoders, these models show good code-generation capability, as we mention some of them in Table \ref{tab:dec-perform}, compared with decoders.

\subsubsection{\textbf{Decoders}}

The decoder-only models are the current SOTA in code generation, as shown in Table \ref{tab:dec-perform}. They have several variants with different architectural nuances. We can broadly categorize them into two - foundational and finetuned.

\begin{table*}[!t]
\centering
\resizebox{\textwidth}{!}{%
\begin{tabular}{@{}c|cccc|cccc|cc@{}}
\toprule
Models  & Architecture & Type & Size & Open Source? & \textbf{HumanEval \cite{black2022gpt}} & \textbf{HumanEval+ \cite{liu2023your}} & \textbf{MBPP \cite{austin2021program}} & \textbf{MBPP+ \cite{liu2023your}} & \textbf{Avg} \\ 
\midrule
Claude 3 [Opus]\cite{anthropic2024claude3} & Decoder & Foundational [Base] & -- & Only Weights & \CellColor{81.7} & \CellColor{75.9} & \CellColor{87.3} & \CellColor{74.1} & \CellColor{81.2} \\
GPT4 Omni\cite{gpt4omni} & Decoder & Foundational [Base] & -- & No & \CellColor{88.4} & \CellColor{81.9} & \CellColor{83.5} & \CellColor{70.8} & \CellColor{81.1} \\
GPT4 Turbo\cite{openai2023gpt4} & Decoder & Foundational [Base] & -- & No & \CellColor{88.4} & \CellColor{81.7} & \CellColor{83} & \CellColor{70.7} & \CellColor{80.9} \\
GPT4\cite{openai2023gpt4} & Decoder & Foundational [Base] & -- & No & \CellColor{85.4} & \CellColor{81.7} & \CellColor{83.5} & \CellColor{70.7} & \CellColor{80.3} \\
Codestral \cite{mistral2} & Decoder & Foundational [Code] & \CellColor{22}B & No & \CellColor{81.9} & \CellColor{76.8} & \CellColor{86.5} & \CellColor{74.2} & \CellColor{80.2} \\
Mistral Large 2 \cite{mistral7b} & Decoder & Foundational [Base] & \CellColor{123}B & No & \CellColor{82.9} & \CellColor{76.8} & \CellColor{86.5} & \CellColor{69.2} & \CellColor{80.3} \\
LLaMA 3 \cite{llama3} & Decoder & Foundational [Base] & \CellColor{405}B & Only Weights & \CellColor{82.7} & \CellColor{76.9} & \CellColor{86.3} & \CellColor{69.1} & \CellColor{80.2} \\
WizardCoder\cite{luo2023wizardcoder} & Decoder & Finetuned & \CellColor{34}B & Only Weights & \CellColor{79.9} & \CellColor{73.2} & \CellColor{78.9} & \CellColor{66.9} & \CellColor{78.9} \\
Nemotron\cite{nemotron} & Decoder & Foundational [Base] & \CellColor{340}B & Only Weights & \CellColor{73.2} & \CellColor{65.9} & \CellColor{82.7} & \CellColor{78.4} & \CellColor{78.5} \\
DBRX\cite{granite} & Decoder & Foundational [Base] & \CellColor{134}B & Only Weights & \CellColor{73.2} & \CellColor{65.9} & \CellColor{82.7} & \CellColor{70.4} & \CellColor{76.4} \\
GPT3.5 Turbo\cite{black2022gpt} & Decoder & Foundational [Base] & \CellColor{175}B & No & \CellColor{73.2} & \CellColor{65.9} & \CellColor{81.7} & \CellColor{69.4} & \CellColor{72.5} \\
MagiCoder\cite{wei2023magicoder} & Decoder & Finetuned & \CellColor{7}B & Weights \& Data & \CellColor{76.8} & \CellColor{70.7} & \CellColor{75.4} & \CellColor{64.4} & \CellColor{71.8} \\
CodeLLaMA\cite{roziere2023code} & Decoder & Foundational [Code] & \CellColor{70}B & Weights \& Data & \CellColor{77.4} & \CellColor{71.3} & \CellColor{75.4} & \CellColor{61.7} & \CellColor{71.5} \\
Phind\cite{phind} & Decoder & Finetuned & \CellColor{34}B & Only Weights & \CellColor{71.3} & \CellColor{67.1} & \CellColor{65.3} & \CellColor{55.1} & \CellColor{64.7} \\
Gemini [Pro]\cite{team2023gemini} & Decoder & Foundational [Base] & -- & No & \CellColor{63.4} & \CellColor{55.5} & \CellColor{72.9} & \CellColor{57.9} & \CellColor{62.4} \\
CodeLLaMA\cite{roziere2023code} & Decoder & Foundational [Code] & \CellColor{34}B & Weights \& Data & \CellColor{51.8} & \CellColor{42.7} & \CellColor{67.2} & \CellColor{52.9} & \CellColor{53.6} \\
phi-1\cite{gunasekar2023textbooks} & Decoder & Foundational [Code] & \CellColor{1.1}B & Only Weights & \CellColor{50.6} & \CellColor{55.5} & \CellColor{53.7}$^{*}$ & \CellColor{52.5}$^{*}$ & \CellColor{53.1} \\
WizardCoder\cite{luo2023wizardcoder} & Decoder & Finetuned & \CellColor{15}B & Only Weights & \CellColor{51.9} & \CellColor{45.1} & \CellColor{61.9} & \CellColor{50.6} & \CellColor{52.4} \\
phi-2\cite{li2023textbooks2} & Decoder & Foundational [Code] & \CellColor{2.7}B & Only Weights & \CellColor{48.2} & \CellColor{43.3} & \CellColor{61.9}$^{*}$ & \CellColor{51.4}$^{*}$ & \CellColor{51.8} \\
StarCoder2\cite{lozhkov2024starcoder} & Decoder & Foundational [Code] & \CellColor{7}B & Weights \& Data & \CellColor{46.3} & \CellColor{37.8} & \CellColor{66.2} & \CellColor{53.1} & \CellColor{50.9} \\
CodeLLaMA\cite{roziere2023code} & Decoder & Foundational [Code] & \CellColor{13}B & Weights \& Data & \CellColor{42.7} & \CellColor{36.6} & \CellColor{61.2} & \CellColor{50.9} & \CellColor{47.9} \\
CodeLLaMA\cite{roziere2023code} & Decoder & Foundational [Both] & \CellColor{7}B & Weights \& Data & \CellColor{34.1} & \CellColor{37.8} & \CellColor{57.6}$^{*}$ & \CellColor{45.4}$^{*}$ & \CellColor{43.7} \\
PaLM-Coder\cite{anil2023palm} & Decoder & Finetuned & \CellColor{540}B & No & \CellColor{35.9} & \CellColor{47.0} & \CellColor{38.1} & \CellColor{44.9} & \CellColor{41.5} \\
StarCoder\cite{li2023starcoder} & Decoder & Foundational [Code] & \CellColor{15}B & Weights \& Data & \CellColor{34.1} & \CellColor{29.3} & \CellColor{55.1} & \CellColor{46.1} & \CellColor{41.2} \\
CodeT5+\cite{wang2023codet5+} & Encoder-Decoder & Foundational [Both] & \CellColor{16}B & Weights \& Data & \CellColor{31.7} & \CellColor{26.2} & \CellColor{54.6} & \CellColor{44.4} & \CellColor{39.2} \\
CodeT5+\cite{wang2023codet5+} & Encoder-Decoder & Foundational [Both] & \CellColor{6}B & Weights \& Data & \CellColor{29.3} & \CellColor{23.8} & \CellColor{51.9} & \CellColor{40.9} & \CellColor{36.5} \\
Mistral\cite{jiang2023mistral} & Decoder & Foundational [Base] & \CellColor{7}B & Weights \& Data & \CellColor{28.7} & \CellColor{23.2} & \CellColor{50.1} & \CellColor{40.9} & \CellColor{35.7} \\
StarCoder\cite{li2023starcoder} & Decoder & Foundational [Code] & \CellColor{7}B & Weights \& Data & \CellColor{24.4} & \CellColor{20.7}$^{*}$ & \CellColor{33.1} & \CellColor{28.8} & \CellColor{26.8} \\
AlphaCode\cite{li2022competition} & Encoder-Decoder & Foundational [Both] & \CellColor{1.1}B & Weights \& Data & \CellColor{17.1} & \CellColor{12.3} & \CellColor{11.8} & \CellColor{21.2} & \CellColor{15.6} \\
SantaCoder\cite{allal2023santacoder} & Decoder & Foundational [Code] & \CellColor{1.1}B & Weights \& Data & \CellColor{14.6} & \CellColor{12.8} & \CellColor{13.8}$^{*}$ & \CellColor{5.6}$^{*}$ & \CellColor{11.7} \\
\bottomrule
\end{tabular}
}
\caption{Performance of Encoder-Decoder and Decoder-only models in Code Generation (\textbf{Percentage in Pass@1}) on several BenchMarks. The larger the model size and the better their performance, the darker their cell color. Also, $^{*}$ denotes the results that are not reported in the original papers and obtained from our own experiments.}
\label{tab:dec-perform}
\end{table*}

\paragraph{\textbf{Foundational}} 
Most notable mentions include the GPT series \cite{radford2018improving} \cite{radford2019language} \cite{brown2020language} \cite{openai2023gpt4}, which evolved from minimal finetuning of GPT1 to zero-shot SOTA results of GPT4. The LLaMA series \cite{touvron2023llama} \cite{touvron2023llama2} also represents major strides through optimized performance and efficiency.

Based on the pretraining procedure, we categorize these foundational models into three:

\begin{enumerate}
    \item \textbf{Base:} These models are pretrained on extensive corpora and designed for general tasks. Large base models like GPT-4 \cite{openai2023gpt4}, Claude 3 \cite{anthropic2024claude3}, Gemini \cite{team2023gemini}, LLaMA (70B) \cite{touvron2023llama2}, and Falcon (180B) \cite{almazrouei2023falcon} excel at code generation, likely due to their substantial parameters and pretraining data, as supported by the Chinchilla Scaling Hypothesis \cite{hoffmann2022training}. This hypothesis posits that performance improves in proportion to the number of parameters and training tokens. In practice, smaller models like LLaMA (7B) \cite{touvron2023llama2}, GPT-2 \cite{radford2019language}, and Falcon (7B) \cite{almazrouei2023falcon} underperform compared to larger ones. However, innovative architectures allow smaller models to remain competitive. The LLaMA authors \cite{touvron2023llama} found their 7B model continued to improve even after training on 1 trillion tokens, challenging conventional scaling laws.

    Unlike GPT models \cite{radford2019language}\cite{black2022gpt}, which use Layer Normalization, LLaMA \cite{touvron2023llama}\cite{touvron2023llama2} employs RMS Normalization for better efficiency. LayerNorm normalizes input across features by adjusting it based on mean and variance, but it can be computationally slow. RMSNorm, on the other hand, normalizes using the root mean square of input values, reducing training time while maintaining performance. Most open-source LLMs use RMSNorm, with Falcon being an exception \cite{almazrouei2023falcon}, as it retains LayerNorm.

    LLaMA models \cite{touvron2023llama2} adopt Rotary Positional Embeddings (RoPE) \cite{su2021roformer} instead of absolute or relative embeddings. Absolute Positional Embeddings \cite{vaswani2017attention}, used in early transformers, encode token positions with sine and cosine functions, while Relative Positional Embeddings \cite{shaw2018self}, used in models like GPT \cite{openai2023gpt4} and Falcon \cite{almazrouei2023falcon}, consider distances between tokens. RoPE combines these, offering improved positional information handling.

    Notably, smaller models like Mistral (7B) outperform larger models like LLaMA (34B) \cite{touvron2023llama2} in code generation, likely due to the Sliding Window Attention mechanism, which enhances long-sequence processing without the quadratic complexity of traditional Causal Masking Attention. This mechanism limits the attention scope to a local window, increasing efficiency, as seen in larger Mistral models \cite{jiang2024mixtral}.
    
    Recent advances in open-source base models include Mistral Large 1 and 2 \cite{mistral2,mistral7b}, DBRX \cite{granite}, Command R+ \cite{alice}, and Nemotron \cite{nemotron}. Mistral Large 2 excels in code generation, supporting over 80 programming and 13 natural languages, outperforming larger models like LLaMA 3.1 (405B) on specific benchmarks. It’s fine-tuned to reduce hallucinations and improve reasoning, making it versatile for developers and researchers.
    
    DBRX \cite{granite} focuses on large-scale code generation and debugging, while Command R+ \cite{alice} enhances real-time coding productivity through suggestions and optimizations. Nemotron \cite{nemotron} specializes in code generation for niche domains like scientific computing and financial modeling, with advanced function-calling capabilities for precise task execution.

    \item \textbf{Code:} General-purpose models like Falcon \cite{almazrouei2023falcon} and LLaMA \cite{touvron2023llama} tend to underperform in code generation due to limited source code in their pretraining corpora. In contrast, models like Code-LLaMA \cite{touvron2023llama}, SantaCoder \cite{allal2023santacoder}, and StarCoder \cite{li2023starcoder} leverage corpora with a higher proportion of source code, such as CodeParrot (1TB) \cite{CodeParrot}, which contains about 80% code and related natural languages. This approach boosts performance. Microsoft's phi-series models \cite{gunasekar2023textbooks}, using a smaller but higher-quality dataset, also outperform models trained on noisier, larger corpora.

    The phi-series utilizes Multi-Head Attention (MHA), which processes information from different subspaces simultaneously. Meanwhile, SantaCoder and StarCoder adopt Multi-Query Attention (MQA), where multiple queries are handled at once, a method also used in GPT-3 \cite{black2022gpt}. Though more efficient, MQA trades off some performance compared to MHA.
    
    \item \textbf{Base + Code:} Larger base models excel in code generation but benefit from additional pretraining on code corpora. LLaMA2 evolves into Code-LLaMA \cite{roziere2023code} after further training on 859 GB of code, while DeepMind’s PaLM-Coder \cite{anil2023palm} adopts a similar approach with a 540B parameter model.
    
    While PaLM-Coder employs MQA, Code-LLaMA introduces Grouped Multi-Query Attention, which processes queries in groups, reducing computational complexity while maintaining efficiency. LLaMA models also optimize resource usage through KV-caching, storing keys and values to reduce computational demand for long sequences, accelerating inference during text generation \cite{jiang2023mistral}.
    
    Both PaLM-Coder and Code-LLaMA implement SwiGLU activations \cite{swishglu}, a blend of Swish \cite{ramachandran2017searching} and GLU \cite{dauphin2017language}, which has proven more effective than traditional ReLU \cite{nair2010rectified} activations.
    
\end{enumerate}

\paragraph{\textbf{Finetuned}} Due to the high costs and resources required for pretraining LLMs, researchers often enhance base or code-pretrained models with supervised finetuning for code generation. Even finetuning smaller LLMs (<7B parameters) demands significant resources, prompting the development of efficient finetuning methods.

One such method is \text{Precision \& Quantization}, which reduces model precision from formats like FP32 to FP16 or INT8, decreasing memory usage and speeding up computation. Quantization further compresses models by discretizing continuous values. However, while improving efficiency, these techniques may reduce model accuracy, necessitating careful trade-offs in resource-constrained environments.

Another approach is \text{LoRA} (Low-Rank Adaptation) \cite{hu2021lora}, which modifies the model's weights by adding low-rank updates without significantly changing the pretrained weights. This method allows efficient adaptation while retaining the learned structure of the base model.

\text{QLoRA} \cite{dettmers2023qlora} extends LoRA by quantizing the low-rank matrices, further reducing the memory footprint and computational resources required for both training and inference, making it especially suitable for deployment in resource-limited settings.

Several works have focused on LLMs finetuned on foundational models for coding tasks. CodeLLama (7B) and CodeLLaMA-Python \cite{roziere2023code} by MetaAI employ a unique approach, using larger models to generate synthetic code corpora for finetuning smaller 7B models. WizardCoder \cite{luo2023wizardcoder} builds on the EvolInstruct method from WizardLM \cite{xu2023wizardlm}, adapting it for coding tasks and finetuning StarCoder \cite{li2023starcoder} with challenging prompts, resulting in models that outperform most open-source code-LLMs on HumanEval, aligning with the Scaling Hypothesis \cite{hoffmann2022training}. OctoCoder \cite{muennighoff2023octopack} also adopts StarCoder \cite{li2023starcoder} as its base, finetuning it on the CommitPack \cite{hoffmann2022training} dataset using LoRA \cite{hu2021lora}, reporting improved results compared to WizardCoder.

Phind-CodeLLaMA \cite{phind} is fine-tuned on a proprietary dataset of high-quality programming problems and solutions, eschewing LoRA \cite{hu2021lora} in its fine-tuning process. MagiCoder \cite{wei2023magicoder}, the smallest open-source model to surpass Code-LLaMA (70B), constructs the OSSInstruct \cite{wei2023magicoder} dataset using GPT3.5 \cite{black2022gpt} to generate higher-quality problem descriptions, then finetunes CodeLLaMA (7B) \cite{roziere2023code} without LoRA \cite{hu2021lora}, achieving a higher HumanEval \cite{black2022gpt} score than GPT3.5.

CodeStral \cite{mistral2}, a 22B model by Mistral AI, supports over 80 programming languages and excels in various code generation tasks, outperforming other models in benchmarks like HumanEval and RepoBench. Lastly, CodeGemma \cite{yadav2024codegemma}, derived from Gemma \cite{gemma2024}, offers 2B and 7B variants with base and instruction-tuned versions, trained on permissively-licensed code and coding-related web pages. It demonstrates strong performance across code-related tasks and employs responsible AI practices.

Most finetuned models tend to focus on the most widely used PLs (Python, JAVA etc.), with limited focus on less used and newly introduced ones. Mojo-Coder \cite{raihan2024mojobench} being the only exception that focuses on the newly introduced Mojo language, mainly used in the AI domain.

\subsection{\textbf{Benchmarks}}
Encoder-only models address diverse coding tasks, while decoder-only and encoder-decoder models focus on generative capabilities. Comparing encoder-only models is challenging due to varied tasks and selective reporting. Code Generation, however, benefits from established benchmarks like HumanEval \cite{black2022gpt} and MBPP \cite{austin2021program}, using the standardized \textit{pass@k} metric \cite{chen2021evaluating}. This metric measures the probability that at least one of k-generated code samples passes all test cases, enabling uniform evaluation of models' task-solving abilities.

\begin{table}[!h]
\centering
\scalebox{0.99}{
\begin{tabular}{@{}llll@{}}
\toprule
Benchmark & PL & NL & Test Cases \\
\midrule
HumanEval & Python & English & 3 per problem \\
HumanEval+ & Python & English & $\sim$240 per problem \\
HumanEval-XL & 12 PLs & 23 NLs & Varies \\
mHumanEval & 25 PLs & 204 NLs & Varies \\
MBPP & Python & English & 3 per problem \\
MBPP+ & Python & English & $\sim$240 per problem \\
Multipl-E & 19 PLs & English & Varies \\
DS-1000 & Python & English & Varies \\
\bottomrule
\end{tabular}
}
\caption{Overview of Code Generation Benchmarks}
\label{tab:code_gen_benchmarks}
\end{table}

The most widely used ones are shown in Table \ref{tab:code_gen_benchmarks}; these provide a comprehensive framework for evaluating code generation capabilities across various programming languages and problem types. HumanEval \cite{black2022gpt} serves as the primary benchmark, with HumanEval+ \cite{liu2023your}, HumanEval-XL \cite{peng2024humaneval} and mHumanEval \cite{raihan2024mhumaneval} offering expanded versions. MBPP \cite{austin2021program} and its extension MBPP+ \cite{liu2023your} focus on Python programming problems. Multipl-E \cite{cassano2022multipl} extends support to multiple programming languages, DS-1000 \cite{lai2023ds} offers diverse, real-world prompts from StackOverflow and CSEPrompts \cite{raihan2024cseprompts,raihan2024performance} focus on prompts from academic settings.

\subsection{\textbf{Challenges}} \label{challenges}

Code LLMs face numerous challenges in development and deployment \cite{she2023pitfalls, liang2022holistic}.

\subsubsection{\textbf{Pitfalls}}
Data quality issues, including sampling bias, noise, and labeling errors, affect model generalization \cite{lu2021codexglue, allal2023santacoder}. Design limitations such as data leakage, spurious correlations, and inappropriate architectures hinder performance \cite{allal2023santacoder, peng2023impact}. Evaluation is complicated by inappropriate baselines, unrealistic benchmarks, and improper metrics \cite{cassano2022multipl, liu2023your}. Deployment challenges include resource constraints, latency requirements, and security threats \cite{she2023pitfalls, pierazzi2020software}. Additionally, models may produce buggy code that is syntactically correct but semantically incorrect or insecure \cite{pearce2022asleep}.

\subsubsection{\textbf{Implications}}
These pitfalls lead to overestimated model performance, difficulty in reproducing results, and vulnerability to attacks such as evasion, data poisoning, and extraction \cite{hendrycks2021unsolved}.

\subsubsection{\textbf{Potential Solutions}}
Improving data quality through initiatives like The Stack \cite{kocetkov2022stack} and phi-series \cite{gunasekar2023textbooks} is crucial. Preventing data leakage and standardizing evaluation methods can enhance result reliability \cite{liang2022holistic}. Incorporating code structures such as syntax trees and control flow graphs may improve model performance \cite{guo2020graphcodebert, wang2021syncobert}. Developing robust models and integrating code analysis tools can address security concerns \cite{pierazzi2020software}.

Our paper contributes by highlighting the need for realistic benchmarks, consistent evaluation metrics, and improved data quality and model robustness.

\section{Open Problems}

While the field of Code LLMs is rapidly evolving, frequently overcoming complex challenges and achieving new state-of-the-art results, several open problems require attention. In this section, we draw conclusions from our survey to highlight these issues, thereby addressing the goals outlined in the introduction.

\subsection{\textbf{Regarding the Benchmarks}}

\paragraph{\textbf{Current benchmarks do not capture real-world scenarios}} Benchmarks like HumanEval \cite{black2022gpt} and MBPP \cite{austin2021program} do not fully capture real-world coding scenarios. There is a need for benchmarks that represent complex multi-function and multi-file problems, tasks with vague specifications, and require documentation skills to better evaluate the models.

\paragraph{\textbf{LLMs need more coverage for a larger range of programming languages}} While most benchmarks are intended for Python, some efforts to broaden them to include more programming languages have been made. Multiple-E \cite{cassano2022multipl} and HumanEval-XL \cite{peng2024humaneval} contain prompts for 18 and 12 programming languages respectively, still missing out on many widely used ones\footnote{\url{https://www.tiobe.com/tiobe-index/}} like Visual Basic, MATLAB, Fortran, etc. These multilingual benchmarks also show a disproportionate focus skewed more towards Python compared to the rest. There are approximately 9,000 programming languages globally\footnote{\url{https://devskiller.com/blog/how-many-programming-languages/}}\footnote{\url{https://www.epitech-it.es/how-many-programming-languages-are/}}, more than the number of natural languages (6,909).\footnote{\url{https://www.linguisticsociety.org/content/how-many-languages-are-there-world}} With at least 250 actively used daily and 100 having a large user base\footnote{\url{https://www.tiobe.com/tiobe-index/}}, there is a clear need for benchmarks that cover more programming languages.

\paragraph{\textbf{Benchmarks may cause LLMs to become biased toward the way they are tested}} While public benchmarks benefit research, they also risk data memorization by allowing Code LLMs to fine-tune on them and achieve misleadingly high results. An ongoing concern in the domain is whether the LLMs are already fine-tuned on the public benchmarks. Introducing benchmarks with blind test sets could help prevent data leakage and model bias toward test data.

\paragraph{\textbf{LLMs are currently limited in their support for the vast majority of human languages}} LLMs are primarily trained on English data, with notable exceptions like GPT-4 \cite{openai2023gpt4}, BLOOM \cite{workshop2022bloom}, and the recent Aya model \cite{ustun2024aya}. However, most LLMs underperform in non-English languages, as indicated by Yiqiao et al. \cite{jin2023better}. Further study is needed to determine if this shortcoming extends to coding tasks as well. It is crucial to know if a Code LLM is as effective when the coding task is described in French or any non-English language.

\subsection{\textbf{Regarding the Models}}

\paragraph{\textbf{Effectiveness of synthetic data suggests quality over quantity}} Recent studies, such as the phi-series \cite{gunasekar2023textbooks,li2023textbooks2} and Magicoder \cite{wei2023magicoder}, have demonstrated the effectiveness of using synthetic datasets for pre-training and fine-tuning. These approaches have yielded impressive results, suggesting that the quality of programming language data may be more crucial than its quantity. Consequently, these findings raise important questions about the future of programming language dataset curation, prompting researchers to reconsider the necessity of large, noisy corpora and explore the potential benefits of focusing on carefully crafted synthetic ones.

\paragraph{\textbf{Need for resource-efficient Code LLMs without performance sacrifice}} LLMs, even smaller (<7B parameters) ones, are computationally very demanding. Adaptation methods like LoRA \cite{hu2021lora} and QLoRA \cite{dettmers2023qlora} help but may compromise performance, requiring further research to minimize this sacrifice. Moreover, the recent success of 1-bit LLMs with BitNet \cite{ma2024era} invites exploration in the context of Code LLMs, potentially significantly reducing resource requirements.

\paragraph{\textbf{Trade-offs in multilingual Code LLMs: Do they lose PL-specific performance?}} The mBERT \cite{devlin2018bert} model, pretrained on 104 natural languages, shows a performance drop for most languages compared to the monolingual BERT models, which is later mitigated by XLM-RoBERTa \cite{conneau2019unsupervised}. As Code LLMs are pretrained and/or fine-tuned on more and more programming languages, it is important to investigate whether they lose any language-specific performance. If so, research is needed on how to reduce this loss.

\paragraph{\textbf{Re-evaluating the Scaling Hypothesis for Code LLMs}} The Chinchilla scaling hypothesis \cite{hoffmann2022training} posits a linear relationship between tokens and parameters, primarily observed in general LLMs. Given the phi-1 model's \cite{gunasekar2023textbooks} success with higher-quality data over sheer size, there is a need to reconsider the scaling hypothesis for Code LLMs with a focus on data quality. Further study is warranted on what constitutes \textit{quality data} in this context.

\paragraph{\textbf{Do Code LLMs perform worse on general tasks?}} Code LLMs are often pretrained on corpora that have significantly more programming language content compared to natural language data. Does this make them perform worse on more general tasks aside from coding tasks? Additionally, do code fine-tuned models become worse than their base versions at regular tasks? These questions require further investigation.

% \subsection{\textbf{Regarding the Use Cases}}

% \paragraph{\textbf{Unexplored prompting techniques and models for Code LLMs}} There have been several works enhancing LLMs' capabilities with innovative and tailored use cases. Prompting techniques such as Chain-of-Thought \cite{wei2022chain}, Tree-of-Thought \cite{yao2024tree}, and Graph-of-Thought \cite{besta2023graph} have proven to be very effective in various cases, yet have not been explored in terms of coding tasks. Retrieval-Augmented Generation (RAG) \cite{lewis2020retrieval} for low-resource programming languages and specific use cases can also be very useful. Lastly, Mixture-of-Experts models, recently popularized by Mixtral \cite{jiang2024mixtral}, are also an unexplored area for Code LLMs.

\section{Conclusion}

This paper presents a comprehensive survey of Code Language Models (Code LLMs), organized around a carefully developed taxonomy that encapsulates the core aspects of the field. By tracing the evolution of Code LLMs, the survey distills the most significant concepts into a clear and structured framework, which is discussed in detail. Each subdomain within Code LLMs is examined thoroughly, identifying current research challenges, gaps in knowledge, and offering insights into potential solutions that could drive future advancements.

Beyond the conceptual analysis, the survey provides an extensive review of existing models, focusing on their architectural intricacies, training methodologies, and fine-tuning techniques. The survey is current, incorporating the most recent models, datasets, and research findings as of August 2024, ensuring its relevance to ongoing developments in the field.

The focus is on works that have been rigorously evaluated, either through publication in well-regarded journals or validation on recognized benchmarks and leaderboards. These benchmarks and leaderboards are maintained by established and influential communities, such as EvalPlus \cite{liu2024your}, AllenAI's WildBench\footnote{\url{https://huggingface.co/spaces/allenai/WildBench}}, \footnote{\url{https://github.com/LudwigStumpp/llm-leaderboard}} and HuggingFace's BigCode LLM\footnote{\url{https://huggingface.co/spaces/bigcode/bigcode-models-leaderboard}}. This approach ensures that the survey captures the most credible and impactful developments in the field.

Overall, this survey aims to serve as a valuable resource for researchers and practitioners in the domain of Code LLMs. By providing a clear understanding of the current state of the art, highlighting open problems, and suggesting directions for future research, it contributes to the advancement of knowledge and practice in this rapidly evolving area.

\section*{Acknowledgments}

We would like to thank the anonymous WLLFM 2024 workshop reviewers for their insightful feedback. 

\bibliographystyle{IEEEtran}
\bibliography{IEEEexample}

\end{document}